\pdfoutput=1
\documentclass{article} 
\usepackage{geometry}

\usepackage{amsthm,amsmath,amssymb}
\usepackage{authblk}
\usepackage{booktabs} 
\usepackage{caption}
\usepackage{graphicx}
\usepackage{subcaption}
\usepackage{xspace}
\usepackage[inline]{enumitem}

\PassOptionsToPackage{numbers}{natbib}
\usepackage{natbib}

\def\BibTeX{{\rm B\kern-.05em{\sc i\kern-.025em b}\kern-.08emT\kern-.1667em\lower.7ex\hbox{E}\kern-.125emX}}

\begin{document}
\title{Constructing High Precision Knowledge Bases\\with Subjective
  and Factual Attributes}

\author{Ari Kobren$^{\dag}$, Pablo Barrio$^{\ddag}$, Oksana Yakhnenko$^{\ddag}$,\\Johann Hibschman$^{\ddag}$,
  Ian Langmore$^{\ddag}$}
\affil{College of Information and Computer Sciences \\ University of
  Massachusetts Amherst$^\dag$}
\affil{Google Inc., New York City$^{\ddag}$}
\affil{akobren@cs.umass.edu, \{pjbarrio,oksana,johannh,langmore\}@google.com}

\maketitle

\newcommand{\leb}{\textsc{ML}\xspace}

\begin{abstract}
  Knowledge bases (KBs) are the backbone of many ubiquitous
  applications and are thus required to exhibit high
  precision. However, for KBs that store \emph{subjective attributes}
  of entities, e.g., whether a movie is \textsf{kid friendly}, simply
  estimating precision is complicated by the inherent ambiguity in
  measuring subjective phenomena. In this work, we develop a method
  for constructing KBs with \emph{tunable precision}--i.e., KBs that
  can be made to operate at a specific false positive rate, despite
  storing both difficult-to-evaluate subjective attributes and more
  traditional factual attributes. The key to our approach is
  probabilistically modeling user \emph{consensus} with respect to
  each entity-attribute pair, rather than modeling each pair as either
  True or False. Uncertainty in the model is explicitly represented
  and used to control the KB's precision. We propose three neural
  networks for fitting the consensus model and evaluate each one on
  data from Google Maps--a large KB of locations and their subjective
  and factual attributes. The results demonstrate that our learned
  models are well-calibrated and thus can successfully be used to
  control the KB's precision. Moreover, when constrained to maintain
  95\% precision, the best consensus model matches the F-score of a
  baseline that models each entity-attribute pair as a binary variable
  and does not support tunable precision. When unconstrained, our
  model dominates the same baseline by 12\% F-score. Finally, we
  perform an empirical analysis of attribute-attribute correlations
  and show that leveraging them effectively contributes to reduced
  uncertainty and better performance in attribute prediction.
\end{abstract}
\section{Introduction}
\label{sec:intro}
Structured knowledge repositories--known as knowledge bases (KBs)--are
the backbone of many high-impact applications and services. For
example: the Netflix\footnote{https://www.netflix.com/} movie
recommendation engine relies on a KB of user-movie-rating triples,
Google Maps\footnote{https://www.google.com/maps/} is built atop a KB
of geographic points of interest and
PubMed\footnote{https://www.ncbi.nlm.nih.gov/pubmed/} offers a handful
of tools that operate on its citation KB of biomedical
research. Wikipedia\footnote{https://www.wikipedia.org/} is both a KB
(with respect to infoboxes) and a service in and of itself and has
even inspired and facilitated the creation of additional KBs like YAGO
and DBPedia \cite{suchanek2007yago, jl_2014/swj_dbpedia}.

In KBs that support real-world decision making, maintaining high
precision is often critical. As an example, consider organizing a
lunch meeting and issuing a KB query for cafes that are \textsf{good
  for groups}.  In the KB's response, it is far better to omit a few
true positives (i.e., cafes that \emph{are} \textsf{good for groups}) than it
is to return \emph{any} false positives (i.e., cafes that \emph{are not}
\textsf{good for groups}), because choosing to visit a false positive
could lead to a highly unproductive meeting.  The importance of high
precision can be even more pronounced for queries about factual
data--like a query for restaurants that are \textsf{wheelchair
  accessible}--where false positives may be inaccessible by the user
issuing the query. Since most KBs are built using noisy automated
methods, special consideration must be paid. Previous work echos this
concern: in addition to employing trained automated components for
data collection and prediction of missing values, systems that build
KBs often turn to humans--largely considered to be more precise than
the automated methods--for validation, writing inference rules,
identifying relevant features, labeling data and even responding to
queries~\cite{carlson2010toward, bennett2007netflix, niu2012deepdive,
  franklin2011crowddb, marcus2011crowdsourced,
  parameswaran2012deco}.

Supporting control over the precision of KB query-responses requires
that each element of a response have an associated probability of
being True, or score, by which it can be filtered.  For KBs that store
\emph{subjective} data (in addition to factual data), supporting such
control is evasive because of the inherent ambiguity in measuring
subjective phenomena. As a concrete example, consider a query for
\textsf{romantic} locations in Paris, France, and deciding whether or
not to include the Pont des Arts--a.k.a., The Love Lock Bridge--in the
response set. The bridge is neither definitively \textsf{romantic} nor
\textsf{unromantic} making it unclear how best to compute the
probability of it \emph{being} \textsf{romantic}, and thus
complicating the decision of whether or not it should be filtered out.

This work develops a method for constructing a KB of entities and
their attributes that offers \emph{tunable precision}--that is, the KB
can be set to run with a particular false positive rate, even when it
stores subjective attributes. To begin, we define the \emph{yes rate}
of an entity-attribute pair to be the fraction of users who believe
that the entity exhibits the attribute, as the number of surveyed
users goes to infinity. If ground-truth yes rates were available, the
attributes of an entity could be determined by thresholding the
corresponding yes rates. For example, when queried for locations that
are \textsf{romantic} a KB might only return locations with yes rates
higher than 0.9. However, ground-truth yes rates are never fully
observed and attempts to survey enough users for empirical yes rate
estimation for every entity-attribute pair are easily stymied by the
scale of most KBs.

Therefore, we propose a hybrid (i.e., human-machine) approach to
constructing KBs, which at its heart employs a probabilistic model for
yes rate estimation.  Our approach begins with crowdsourcing: we serve
users questions of the form: ``does entity $e$ exhibit attribute
$a$?'' and we receive \emph{yes votes} and \emph{no votes} in response
(users may also abstain).  We use the votes to bootstrap training of a
probabilistic yes rate model for each entity-attribute
pair. Uncertainty in each model is explicitly represented via a
distinct prior distribution. The priors allow for quantifiable
confidence when many votes are available and, in the more common case,
quantifiable uncertainty when votes are scarce. Representing
uncertainty is a crucial component of our approach because it is used
to control the precision of the KB.  When the KB is queried,
entity-attribute pairs are only included in the response if the KB is
sufficiently confident that their corresponding yes rate exceeds the
threshold. As long as the learned models are well-calibrated, this
approach can be used to control the KB's false positive rate. The
procedure can even be used with respect to factual attributes, where
we expect yes rates to be close to 0 or 1 and where representing
uncertainty helps make the KB robust to noise from
crowdsourcing~\cite{sheng2008get}.

We study the KB that supports Google Maps.  This KB stores real-world
landmarks--called \emph{locations}--and their subjective and factual
\emph{attributes}. Since the number of location-attribute pairs in
Google Maps is large, fitting yes rate models using the votes alone
renders most of the models highly uncertain. To mitigate uncertainty,
we leverage \emph{side information} that accompanies each
location--like natural language text extracted from the location's
homepage--during learning. Intuitively, the side information is likely
to be indicative of the location's attributes. For example, finding
the phrase ``wine list'' on a restaurant's homepage may constitute
strong evidence that the restaurant exhibits the attribute,
\textsf{serves alcohol}. To further reduce model uncertainty, we also
promote information sharing across attributes. This is beneficial when
attributes are related. For example, a location that has many yes
votes for the attribute \textsf{romantic} is unlikely to receive many
yes votes for the attribute \textsf{kid friendly}. Both the side
information and shared information can be used to address the
\emph{cold start} problem--i.e., predicting the attributes of a
location with no observed votes. This is critical for large KBs and
for KBs with ever-expanding lists of entities and attributes, like Google
Maps, in which cold starting is common.

We present three neural networks for estimating the yes rate of each
location-attribute pair. Each network uses side information and is
trained using a different style of information sharing across
attributes.  We evaluate the three networks on their ability to
accurately represent uncertainty and predict attributes of the
locations in Google Maps. When constrained to operate with 95\%
precision, our best model improves on the precision of an
unconstrained baseline by 6\% and matches the baseline's F-score. This
also amounts to a 5\% increase in F-score over a neural baseline that
uses multi-task learning--a common paradigm for information
sharing--and that is subject to the same precision constraint (i.e.,
95\%). When unconstrained, our best model dominates both the empirical
and neural baselines by 12\% and 17\% F-score,
respectively. Additionally, our results reveal that some styles of
information sharing lead to improved F-score by bolstering model
confidence while others do not. This observation suggests that
information sharing can be detrimental when performed between two
attributes, one of which is data rich and the other is data
poor. Finally, we demonstrate that our learned models are
well-calibrated via Q-Q plots. While we study the location-attribute
setting, our yes rate modeling framework can be applied in many
instances of hybrid KB construction that rely on collecting
categorical observations via crowdsourcing.
\section{Locations, Attributes and Votes}
\label{sec:background}
We study the problem of constructing a knowledge base (KB) of
locations and their attributes. The term \emph{location} refers to a
real-world landmark (e.g., a restaurant, monument, museum, business,
park, etc.)  and the term \emph{attribute} refers to a characteristic
of a landmark. \emph{Constructing the KB} roughly refers to
determining, for each location-attribute pair, whether the location
exhibits the attribute. For example, the KB should store whether
\textsf{Starbucks at 1912 Pike Pl, Seattle} is a \textsf{local
  favorite}. The KB stores subjective attributes (like \textsf{local
  favorite}) and factual attributes (like \textsf{has free wifi}). A
subset of the attributes can be found in Table
\ref{tab:attributes}. In this work, we focus on the KB underlying
Google Maps, which includes more than 50 million locations and more
than 70 attributes.

\begin{table}[t!]
  \centering
  \begin{tabular}{l l }
    \hline
    Factual & Subjective\\
    \hline
    \texttt{CASH\_ONLY}             & \texttt{BUSTLING} \\
    \texttt{WHEELCHAIR\_ACCESSIBLE} & \texttt{GOOD\_VIEW} \\
    \texttt{ACCEPTS\_RESERVATIONS}  & \texttt{COZY} \\
    \texttt{HAS\_HIGH\_CHAIRS}      & \texttt{KID\_FRIENDLY} \\
    \texttt{SERVES\_FOOD\_LATE}     & \texttt{QUICK\_VISIT}\\
    \hline
  \end{tabular}
  \caption{A sample of factual and subjective attributes.}
  \label{tab:attributes}
  \vskip -0.2in
\end{table}

For each location, the KB has access to associated structured and
unstructured meta-data known as side information. The unstructured
side information for a location is comprised of text extracted from
that location's homepage and from other relevant web pages. The
structured side information includes tags for that location that come
from a proprietary ontology. Both the structured and unstructured side
information are available as natural language text.

\begin{figure}[t!]
  \centering
  \centerline{\includegraphics[width=0.75\columnwidth]{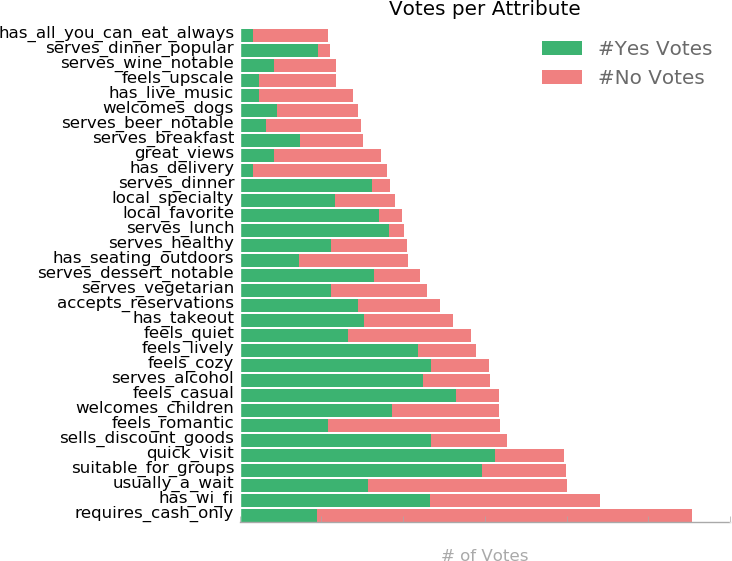}}
  \caption{Yes and no votes for a subset of attributes. Precise
    vote counts have been omitted to respect data sensitivity.}
    \label{fig:votes}
\end{figure}

Crowdsourcing is used to gather \emph{yes votes} and \emph{no votes}
for the location-attribute pairs. Specifically, users who have
recently visited a particular location are served yes-or-no questions
regarding the attributes of that location. Users may supply a yes
vote, a no vote or an abstention in response. Despite the continuous
deployment of these questions, in comparison to the number of
location-attribute pairs, the number of votes is small: there are
\textasciitilde{}2.65 $\times$ $10^{8}$ location-attribute pairs, only
\textasciitilde{}13\% of which have at least 1 associated vote, and,
of those pairs, \textasciitilde{}50\% have only 1 vote.  The votes are
neither evenly distributed among locations or attributes nor are they
guaranteed to be unanimous (i.e., many pairs receive both yes and no
votes). See Figure \ref{fig:votes} for a sketch of the number of yes
and no votes collected for a subset of the attributes.

\section{The Observation Model}
\label{sec:model}
In constructing the KB, it is tempting to assume that each
location-attribute pair is either True or False. However, this
assumption does not hold for many pairs, especially those that include
a subjective attribute. Therefore, we model the consensus among users
with respect to each pair.  In this section, we formalize this notion
of consensus via a generative model of the observed votes.

\subsection{Location-attribute Yes Rate}
\label{subsec:yesrate}
We assume that each location-attribute pair has a latent \emph{yes
  rate}. The yes rate represents the fraction of users who agree that
the location exhibits the attribute, i.e., it is a measure of
consensus. Formally, for location $l$ and attribute $a$ let $0
\le \theta_{la} \le 1.0$ be the yes rate for the pair. We model each
vote for the pair $(l, a)$ as a sample from a pair-specific binomial
distribution. The success parameter of this distribution is equal to
the pair's latent yes rate.  Therefore,
\begin{align*}
  Y_{la} \sim B(k, \theta_{la})
\end{align*}
where $B(\cdot, \cdot)$ is the binomial distribution, $\theta_{la}$ is
the yes rate and $Y_{la}$ is the number of yes votes out of
$k$ total votes for the pair $(l,a)$.

\subsection{Representing Uncertainty}
Ideally, there would be enough votes to reliably estimate the yes rate
for each location-attribute pair. However, the KB contains hundreds of
millions of pairs, most of which have no corresponding
votes. Therefore, we explicitly represent uncertainty in every pair's
yes rate through a pair-specific prior distribution. Specifically,
each $\theta_{la}$ is modeled as a beta distributed random
variable\footnote{The beta distribution is often parameterized by two
  hyperparameters: $\alpha$ and $\beta$. In our equivalent
  parameterization, $\mu = \frac{\alpha}{\alpha + \beta}$ and $\tau =
  \alpha + \beta$. Our parameterization makes the parameters easier to
  interpret.}:
\begin{align*}
  \theta_{la} \sim Beta(\mu_{la}\tau_{la}, \tau_{la}(1 - \mu_{la})).
\end{align*}
Here, $\mu_{la}$ is the \emph{expected yes rate} for the pair
$(l,a)$--that is, $\mu_{la}$ represents the fraction of yes votes as
the total votes goes to infinity:
\begin{align*}
  \mu_{la} = \mathbb{E}\left[\frac{Y_{la}}{Y_{la} + N_{la}}\right]
\end{align*}
where, $Y_{la}$ and $N_{la}$ are the number of yes and no votes for
the pair $(l,a)$, respectively. $\tau_{la}$ is the prior
distribution's \emph{precision}:
in a sense, $\tau_{la}$ captures certainty in the true yes rate being
close to $\mu_{la}$.  This choice in representation precisely defines
KB construction: estimate both $\mu_{la}$ and $\tau_{la}$ for each
location-attribute pair.

There are a number of advantages to this hierarchical beta-binomial
model for the observed votes. Most important for high precision KBs is
that the model facilitates closed-form computation of its
\emph{confidence} in each estimated expected yes rate (further
discussion appears in Section \ref{sec:setup}). As long as the model
is well-calibrated, this allows the KB maintainer to control the false
positive rate by filtering results by model confidence. Specifically,
consider a query, $q$, for which the optimal result set includes all
location-attribute pairs that satisfy $\theta_{la} > \mu_{min}$. To
guarantee a false positive rate that is less than $\delta$ in
expectation, for $q$, the KB maintainer can: gather all pairs where
$\mu_{la} > \mu_{min}$ and filter out all pairs for which the KB has
confidence less than $1-\delta$.

The structure of our model has other practical advantages, too. First,
the prior beta distribution makes yes rate estimation more robust when
only a few (or zero) votes are observed for a pair. Another practical
advantage is that our model structure facilitates efficient updates:
because of conjugacy, after observing additional votes for some
location-attribute pair, it is possible to update the model's
estimates for the corresponding $\mu$ and $\tau$ efficiently. This is
particularly valuable because the KB's size makes fitting the model
very expensive and because new votes are observed frequently.
\section{Attribute Relatedness}
\label{sec:empirical}
\begin{figure}[t!]
  \centering
  \begin{subfigure}[t]{0.75\textwidth}
    \centerline{\includegraphics[height=0.7\columnwidth]{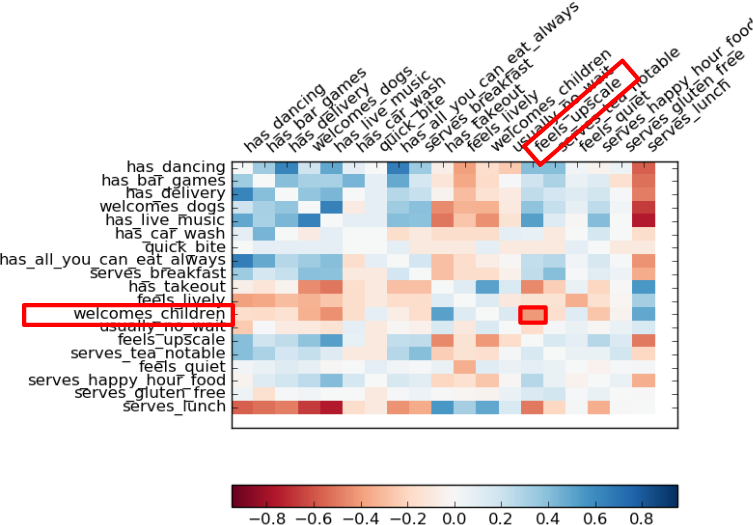}}
    \caption{Attribute pairs that generally exhibit negative
      relatedness.}
    \label{fig:relatedness1}
  \end{subfigure}\\
  \begin{subfigure}[t]{0.75\textwidth}
    \centerline{\includegraphics[height=0.7\columnwidth]{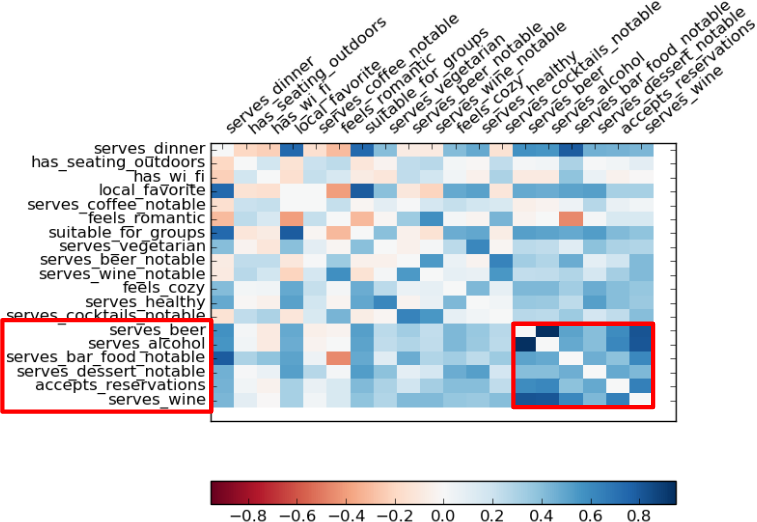}}
    \caption{Attribute pairs that generally exhibit positive
      relatedness.}
    \label{fig:relatedness2}
  \end{subfigure}
  \caption{Submatrices of the relatedness
    matrix. \textsf{welcomes children} and \textsf{feels upscale}
    are negatively related; \textsf{serves beer},
    \textsf{serves alcohol}, \textsf{serves bar food notable},
    \textsf{serves dessert notable}, \textsf{accepts reservation}
    and \textsf{serves wine} are all positively related.}
  \label{fig:relatedness}
\end{figure}
 The parameters of the observation model
(Section \ref{sec:model}) can be learned independently for each
location-attribute pair. But, intuitively, many of the attributes are
closely related to one another. For example, a location that
\textsf{has takeout} probably does not \textsf{feel upscale}. Were
attributes correlated with one another, jointly learning parameters
across attributes would yield more accurate and confident models.

We present a qualitative analysis that highlights strong correlations
between various attribute pairs.  For each attribute-attribute pair,
we count the number of locations for which both attributes
\emph{agree}--i.e., both have majority yes votes or both have majority
no votes--and the number of locations for which the attributes
\emph{disagree}--i.e., one attribute has majority yes votes while the
other has majority no votes.  If there are an equal number of yes and
no votes for a particular location-attribute pair, the attribute
neither agrees nor disagrees with any other attribute (for that
location). We construct a \emph{relatedness} matrix of dimension
$|A|\times|A|$, where $|A|$ is the number of attributes.  Each cell in
the matrix is computed by:
\begin{align*}
  r_{ij} = \frac{\sum_{l \in L}[\mathrm{\emph{agree}}_{ij}] -
    [\mathrm{\emph{disagree}}_{ij}]}{|L| + b}
\end{align*}
where $L$ is the set of all locations and $b$ is a bias term.

Figure \ref{fig:relatedness} shows two quadrants of the relatedness
matrix with rows and columns sorted by increasing sum-total
relatedness and diagonal set to zero.  The relationships depicted
generally match intuition about the attributes that are positively and
negatively related. However, not all pairs respect intuition, largely
due to vote scarcity. For example, the attribute \textsf{has car wash}
has very few votes and exhibits scores that may misrepresent its
relatedness to other attributes.  Regardless of the noise, the
relatedness matrix suggests that learning from attribute-attribute
correlations is useful.

\section{Learning Model Parameters}
\label{sec:learning}
In this section we propose three neural network architectures for
estimating the parameters of the observation model (Section
\ref{sec:model}). Each architecture is designed to leverage
information sharing across attributes differently.

\subsection{Multi-task Baseline}
\label{subsec:baseline}
The multi-task learning baseline architecture (\leb), models each
vote-generating distribution (i.e., $\mu_{la}$ and $\tau_{la}$) as
follows:
\begin{align*}
  \mu_{la} &= \sigma\left(f_{\mu_a}^{(ml)}(e_l)\right)\\
  \tau_{la} &= \log\left[1 + \exp{\left(f_{\tau_a}^{(ml)}(e_l)\right)}\right]
\end{align*}
where $e_l$ is an \emph{embedded representation} of location $l$,
$f_{\mu_a}^{(ml)}(\cdot)$ and $f_{\tau_a}^{(ml)}(\cdot)$ are learnable,
attribute-specific functions of the location embedding and
$\sigma(\cdot)$ is the softmax function. $f_{\mu_a}^{(ml)}(\cdot)$ and
$f_{\tau_a}^{(ml)}(\cdot)$ are trained to facilitate estimation of
$\mu$ and $\tau$, respectively. The superscript $ml$ denotes that these
functions are part of the \leb baseline architecture.  Notice that the
architecture ensures that both $\mu$ and $\tau$ take values in
appropriate domains.

The location embedding and the functions $f_{\mu_a}^{(ml)}(\cdot)$ and
$f_{\tau_a}^{(ml)}(\cdot)$ are trained by maximizing the marginal
likelihood, or \emph{evidence}.  That is, the training objective is to
maximize the likelihood of the observed votes under the beta-binomial
observation model.  Because the beta distribution is conjugate to the
binomial, it is possible to compute the evidence exactly. Letting
$\alpha = \mu_{la}\tau_{la}$ and $\beta = \tau_{la}(1 - \mu_{la})$,
the evidence corresponding to the pair $(l,a)$ is:
\begin{align*}
  \label{eq:like}
    \mathcal{L}_{la}(Y_{la}; k, \mu_{la}, \tau_{la})  =\binom{k}{Y_{la}}\frac{1}{Z(\alpha, \beta)}& \cdot Z(Y_{la} +
  \alpha, k - Y_{la} + \beta)
\end{align*}
where $Z(\alpha, \beta) = \frac{\Gamma(\alpha)\Gamma(\beta)}{\Gamma(\alpha +
  \beta)}$ is the normalizer of the beta distribution.

The \leb model uses \emph{multi-task learning}, a standard technique
for sharing information across related
tasks~\cite{caruana1998multitask}. In practice, the location embedding
is updated with gradients computed with respect to each
attribute. Intuitively, each embedding captures the salient features
of the corresponding location.

\subsection{Alien Vote Architecture}
\label{subsec:av}
One shortcoming of the \leb architecture is that information is shared
indirectly via gradients.  Our second architecture shares information
more directly. In particular, this architecture leverages \emph{alien
  votes}, where the alien votes for a pair $(l,a)$ are all votes for
the attributes of $l$ excluding $a$. Formally, for a pair $(l,a)$, the
alien votes are:
\begin{align*}
  V_{l\bar{a}} = \{(l',a'): l' = l, a' \ne a\}.
\end{align*}
Direct access to alien votes makes it easier to learn relationships
like: a restaurant that has many yes votes for the attribute
$\textsf{romantic}$ is unlikely to be $\textsf{kid friendly}$.

We introduce the following alien vote architecture (AV):
\begin{align*}
  \mu_{la} &= \sigma\left(f_{\mu_a}^{(av)}(e_l \oplus g(V_{l\bar{a}}))\right)\\
  \tau_{la} &= \log\left[1 + \exp{\left(f_{\tau_a}^{(av)}(e_l \oplus g(V_{l\bar{a}}))\right)}\right]
\end{align*}
where the $\oplus$ operator denotes vector concatenation. In this
architecture, the alien votes for a pair $(l,a)$ are transformed using
a learned function $g(\cdot)$ and concatenated with the location
embedding.  The concatenation is then used to compute $\mu_{la}$ and
$\tau_{la}$, facilitating direct learning of attribute-attribute
correlations.

\subsection{Independent AV Architecture}
\label{subsec:iav}
For multi-task learning to be effective, the same features must be
useful for each task.
Since some attributes are unrelated (Section \ref{sec:empirical}), we
introduce an \emph{independent} alien vote (IAV) architecture that
does not employ multi-task learning. Unlike the other architectures,
the IAV architecture uses a separate network per attribute:
\begin{align*}
  \mu_{la} &= \sigma\left(f_{\mu_a}^{(iav)}(e_{la}^+)\right)\\
  \tau_{la} &= \log\left[1 + \exp{\left(f_{\tau_a}^{(iav)}(e_{la}^+)\right)}\right]
\end{align*}
where $e_{la}^+$ is a location embedding that is computed, in part,
from the alien votes $V_{l\bar{a}}$. Thus, the IAV model still uses
attribute-attribute correlations learned from the alien votes.

Under the IAV architecture, attributes with many votes enjoy exclusive
access to the location embeddings and do not suffer from spurious
gradients computed with respect to unrelated attributes. On the other
hand, attributes with few votes cannot leverage the salient location
features learned through multi-tasking.

\section{Implementation}
\label{subsec:arch}
Each architecture is implemented as a feed-forward neural network. The
networks use side information--and optionally alien votes--to estimate the
expected yes rate (and associated uncertainty) of each
location-attribute pair, including the pairs with no observed votes.

\subsection{\leb Baseline}
The input to the \leb architecture is the side information $x_l$ for a
specific location $l$ (Section \ref{sec:background}). We represent
each token in the side information by a 285-dimensional
embedding\footnote{Dimensionality of these embeddings is chosen
  heuristically.}. These tokens are combined by summing their
embeddings and computing the element-wise square root.\footnote{The
  square root is a heuristic that intuitively allows the input vector
  for a location with a large amount of side information to have a
  larger magnitude while not becoming too large.} This 285-dimensional
embedding is passed through 5 fully-connected layers, each with 500
ReLU activation units. The resulting representation of the side
information is the location embedding, $e_l$.

\begin{figure*}[t!]
  \centering
  \begin{subfigure}[t]{0.33\textwidth}
    \centerline{\includegraphics[width=\columnwidth]{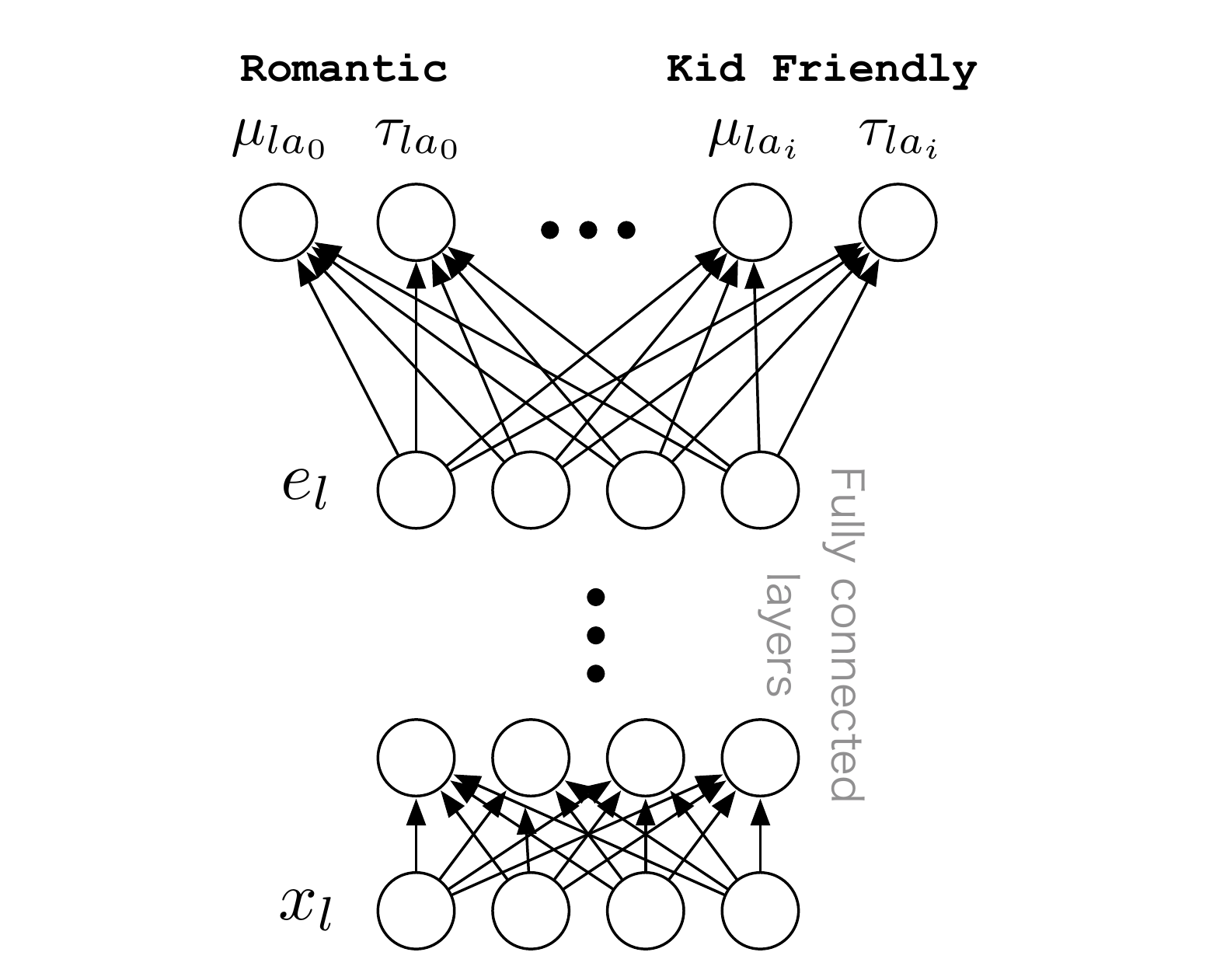}}
    \caption{Location Embedding.}
    \label{fig:baseline}
  \end{subfigure}%
  \begin{subfigure}[t]{0.33\textwidth}
    \centerline{\includegraphics[width=\columnwidth]{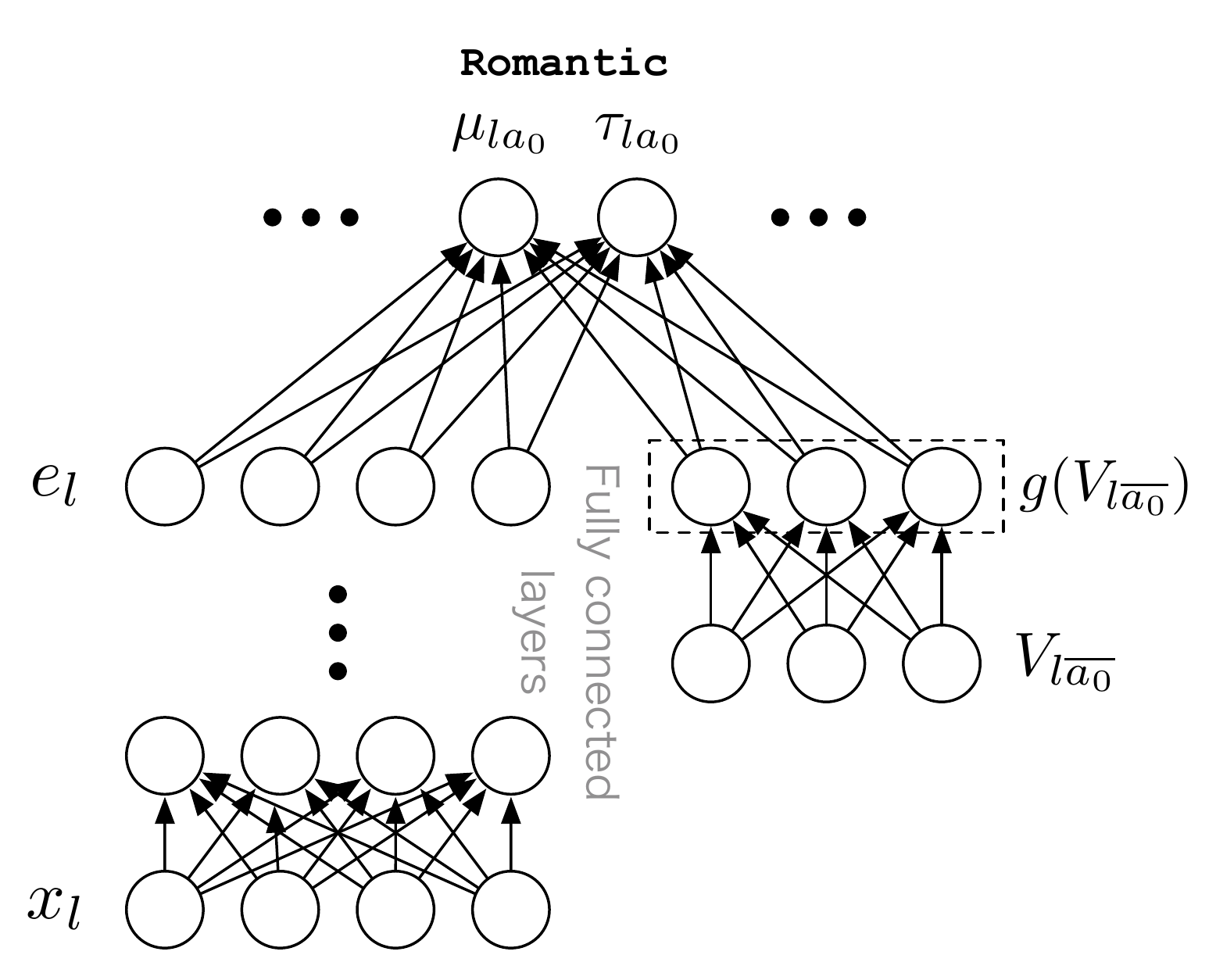}}
    \caption{Alien Votes.}
    \label{fig:av-arch}
  \end{subfigure}%
  \begin{subfigure}[t]{0.33\textwidth}
    \centerline{\includegraphics[width=\columnwidth]{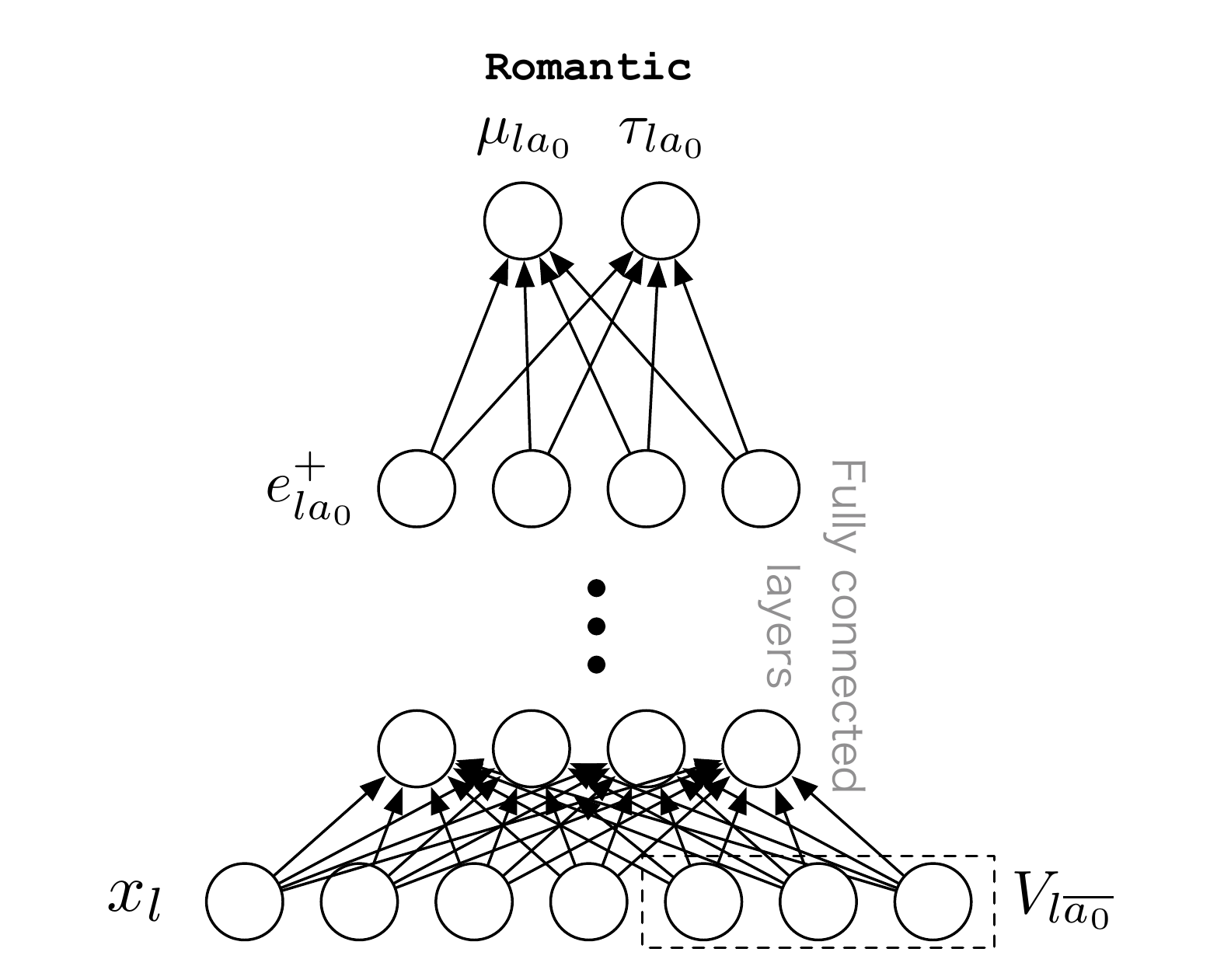}}
    \caption{Independent Alien Votes.}
    \label{fig:iav-arch}
  \end{subfigure}%
  \caption{Three architectures that are trained to estimate parameters
    of the prior distributions of location-attribute
    pairs. Architectures \ref{fig:baseline} and \ref{fig:av-arch} transform
    side information, $x_l$, into a location embedding, $e_l$, which
    can be transformed to estimate the parameters of each attribute
    for location $l$. Architecture \ref{fig:av-arch} concatenates the same
    location embedding, $e_l$, with the transformed alien votes for
    each attribute, $V_{l\bar{a}}$, before estimating the
    corresponding parameters. In architecture \ref{fig:iav-arch}, $x_l$ and
    the $V_{l\bar{a}}$ are transformed into an augmented embedding,
    $e_{la}^+$. Architecture \ref{fig:iav-arch} is duplicated for each
    attribute. In all architectures, the functions
    $f_{\mu_{a}}(\cdot)$ and $f_{\tau_{a}}(\cdot)$ correspond to the
    transformation of the location embedding into $\mu_{la}$ and
    $\tau_{la}$ respectively.}
  \label{fig:arch}
\end{figure*}

The location embedding, $e_l$, is used to estimate $\mu_{la}$ and
$\tau_{la}$ for each attribute. Each attribute $a$ has two parallel
sets of layers, one corresponding to $f_{\mu_a}^{(ml)}(\cdot)$ and the
other corresponding to $f_{\tau_a}^{(ml)}(\cdot)$.  The location
embedding layer is fully connected to both.

To implement multi-task learning (and achieve additional
regularization~\cite{bansal2016ask}) in each training mini-batch, all
gradients computed with respect to $\mu_{la}$ are back-propagated
through $f_{\mu_a}^{(ml)}(\cdot)$ and the location embedding; all
gradients computed with respect to $\tau_{la}$ are back-propagated
through $f_{\tau_a}^{(ml)}(\cdot)$ and also the location embedding. A
visual representation of the network can be found in Figure
\ref{fig:baseline}.

\subsection{AV Architecture}
The implementation of the AV architecture builds on the baseline. In
the AV architecture, the alien votes, $V_{l\bar{a}}$, represented as a
dense vector, are passed through a single fully-connected layer (i.e.,
the function $g(\cdot)$ in Section \ref{subsec:av}) and then
concatenated with the location embedding. While the alien votes may
yield greater predictive power, they also result in an increase in
model parameters and training time. The resulting architecture is
depicted in Figure \ref{fig:av-arch}.

\subsection{IAV Model}
To implement the IAV architecture, we train a separate network for
each attribute (Figure \ref{fig:iav-arch}). In each network, the
alien votes are concatenated with the input rather than the location
embedding. The IAV architecture has more parameters than the AV
architecture but also avoids issues stemming from the multi-task
learning.

\section{Experimental Setup}
\label{sec:setup}
Since the ground-truth yes rate of each location-attribute pair is not
observed, we cannot evaluate a learned model using the residual
between estimated and the ground-truth yes rates. Therefore, we
evaluate the trained models via two other methods. First, we compute
their F-scores in attribute prediction with respect to a set of
\emph{gold labels} (for a subset of locations and their subjective and
factual attributes). Second, we measure model calibration.

\subsection{Model-based Attribute Predictor}
The primary responsibility of the KB is to retrieve all locations that
exhibit a queried attribute. In doing so, recall that it is crucial
for the result set to have very few false positives (while maintaining
as high recall as possible). To accomplish this goal, for each query, the
KB maintainer sets a yes rate threshold, $\mu_{min}$, and a false
positive rate, $\delta$.  For each query, we use a predictor
$s(\cdot, \cdot)$ to build the result set, where the predictor is
defined as:

\[
s(l,a) =
\begin{cases}
  1 & \text{if } Pr(R \ge \mu_{min}; \mu_{la}, \tau_{la}) \ge 1 - \delta\\
  0 & \text{if } Pr(R \le 1-\mu_{min}; \mu_{la}, \tau_{la}) \ge 1 - \delta\\
  \text{No Prediction} & \text{otherwise}
\end{cases}
\]and where $R$ represents a yes rate.  In words, if, with probability at
least $1 - \delta$, the yes rate for a pair $(l,a)$ is greater than
$\mu_{min}$, then the predictor outputs a 1; if, with probability at
least $1 - \delta$, the yes rate for a pair $(l,a)$ is less than $1 -
\mu_{min}$, then the predictor outputs a 0; otherwise there is
insufficient confidence in the yes rate being greater than $\mu_{min}$
or less than $1 - \mu_{min}$ so the predictor outputs ``No
Prediction''. The probabilities used by the predictor are computed via
the cumulative distribution function (CDF) of a beta distribution, for
example:
\begin{align*}
  Pr(R \ge \mu_{min}; \mu_{la}, \tau_{la}) &= \int_{\mu_{min}}^{1}
  Pr(\theta_{la} | \mu_{la}, \tau_{la})d\theta_{la}\\ &= 1 -
  \int_{0}^{\mu_{min}} Pr(\theta_{la} | \mu_{la},
  \tau_{la})d\theta_{la}.
\end{align*}
In our experiments, we use the implementation of the CDF of the beta
distribution included in Tensorflow \cite{tensorflow2015-whitepaper}
and we set the yes rate and uncertainty thresholds as follows:
$\mu_{min} = 0.66$ and $1 - \delta = 0.95$.

\subsection{Data and Evaluation Metrics}
\label{subsec:datafscore}
We evaluate our models on a dataset that includes the real-world
locations and attributes in Google Maps.  The yes and no votes are
collected by soliciting real users for their opinions about the
attributes of the locations they have visited. We collect a separate
set of \emph{gold labels}, $\mathcal{G}$, from vetted workers. The
gold label for the pair $(l,a)$ is a judgment $g_{la} \in \{0, 1\}$,
computed by a majority vote among three vetted workers regarding
whether location $l$ exhibits attribute $a$. For any
location-attribute pair there can be at most 1 corresponding gold
label.

We measure the \emph{F-score} of the attribute predictor $s(\cdot,
\cdot)$ with respect to $\mathcal{G}$. The F-score of the predictor is
the harmonic mean between the predictor's precision and recall. We
report the F-score of the predictor with respect to both the
\emph{prior} and \emph{posterior} parameters (Section \ref{sec:model}).
Because of our model's structure, the posterior parameters
can be computed in closed form:

\begin{align*}
Pr(\theta_{la}|Y_{la}, k) &= \frac{Pr(Y_{la}| k, \theta_{la})Pr(\theta_{la}|\mu_{la},
  \tau_{la})}{\int_{0}^{1}Pr(Y_{la}| k,
  \hat{\theta}_{la})Pr(\hat{\theta}_{la}|\mu_{la},
  \tau_{la})d\hat{\theta}_{la}}\\
 &=Beta(Y_{la} + \mu_{la}\tau_{la},k - Y_{la} + \tau_{la}(1-\mu_{la}))
\end{align*}
where $Z(\cdot, \cdot)$ denotes the beta function (Section
\ref{subsec:baseline}). Note that computing the posterior from the
prior is efficient because it only requires incrementing the
parameters of a beta distribution.  Also, note that increasing $\tau$,
which is related to model confidence, has a similar effect on the
posterior as observing additional votes.

While the F-score of the predictor is a good indication of model
quality, it may be imprecise. To see why, consider the
location-attribute pairs that have a ground-truth yes rate of 0.66,
i.e., $\mathcal{P}_{0.66} = \{(l,a) : \theta_{la} = 0.66\}$. In
expectation, 66\% of votes for these pairs will be yes votes and 34\%
will be no votes (thus reflecting the True yes rate). For each pair in
$\mathcal{P}_{0.66}$ the probability that the gold label will be ``0''
is equal to the probability of at least 2 out of 3 vetted workers
labeling that pair with a ``0'':
\begin{align*}
  Pr(g_{la} = 0 | (l,a) \in \mathcal{P}_{0.66}) &= \binom{3}{2} \cdot 0.66
  \cdot 0.34^{2} + 0.34^{3} = 0.268.
\end{align*}

If a model correctly estimates a yes rate of 0.66 for each pair
$(l,a) \in \mathcal{P}_{0.66}$ with high confidence, then for each
such pair the predictor $s(\cdot,\cdot)$ will output ``1''. Since the
vetted workers only label \textasciitilde{}75\% of these pairs with
``1'', the maximum attainable precision is \textasciitilde{}75\% even
though the model is perfect.

Despite this limitation, we argue that our evaluation scheme
sufficiently captures model quality. First, for attributes with yes
rates close to 0 or 1 (e.g., most factual attributes) the effect of
this imprecision is minor. Second, the gold labels in $\mathcal{G}$
correspond to pairs for which there is high worker agreement (e.g.,
for most pairs, worker votes are unanimous).  Assuming that workers
are reliable, the number of location-attribute pairs with little
consensus in the gold set will be small. We also note that for the
subjective attributes, our evaluation scheme produces a conservative
estimate of model quality, which, we argue, is better than a
non-conservative estimate given the importance of
mitigating false positives.

We supplement the F-score evaluation with model calibration
measurements. These measurements directly evaluate the model's yes
rate estimates rather than the discrete output of the predictor.
\section{Experiments}
For training, we use the Adagrad~\cite{duchi2011adaptive} optimizer with learning
rate of 0.1 and a batch size of 256. We compare the 3 architectures
(Section \ref{sec:learning}) with two additional empirical baselines.
We also analyze how different information sharing paradigms affect
attribute prediction. Finally, we provide Q-Q plots showing that our
trained models are well-calibrated.

\label{sec:experiments}
\subsection{Attribute Prediction}
\label{subsec:av-mech}
We compare the models learned via the 3 architectures. For the AV and
IAV architectures, we test 3 alien vote representations (Section
\ref{subsec:av}):
\begin{enumerate*}
\item \textbf{raw}: the raw counts of both yes and no votes,
\item \textbf{maj}: the majority vote (either yes or no),
\item \textbf{prob}: the expected value and 1 - the expected value of
  a beta distribution (with uniform prior) that is fit using the
  observed votes.
\end{enumerate*}
We compare the models to two empirical baselines. One baseline
(Empirical) predicts a ``1'' when the observed yes rate for a pair is
greater than $\mu_{min}$.  The second, more precise baseline
(Empirical-P) only makes predictions for pairs with at least 3
observed votes. We also show the performance of the IAV raw model
operating in ``high-recall mode'' (IAV-HR), meaning that a ``1'' is
predicted when $\mu_{la}>0.66$ and no additional filtering (based on
confidence) is performed.

\begin{figure}[t!]
  \centering
  \begin{tabular}{ l c c c c c c }
    \hline
    & \multicolumn{3}{c}{Prior} & \multicolumn{3}{c}{Posterior}\\
    Model         & PRE  & REC  & F1   & PRE  & REC  & F1\\
    \hline
    \leb Baseline & \textbf{0.98} & 0.40 & 0.57 & \textbf{0.97} & 0.52 & 0.67\\
    AV raw        & 0.96 & 0.46 & 0.63 & 0.96 & 0.54 & 0.69\\
    AV maj        & 0.95 & 0.42 & 0.58 & 0.95 & 0.52 & 0.68\\
    AV prob       & 0.95 & 0.47 & 0.63 & 0.95 & 0.56 & 0.70\\
    IAV raw       & 0.94 & \textbf{0.51} & \textbf{0.66} & 0.94 & \textbf{0.59} & \textbf{0.72}\\
    IAV maj       & 0.95 & 0.47 & 0.63 & 0.95 & 0.55 & 0.70\\
    IAV prob      & 0.93 & 0.49 & 0.64 & 0.94 & 0.54 & 0.69\\
    \hline
    Empirical     & ---  & ---  & ---    & 0.88 & 0.61 & 0.72 \\
    Empirical-P   & ---  & ---  & ---    & 0.90 & 0.41 & 0.56 \\
    IAV-HR        & 0.88 & 0.77 & 0.82   & 0.89 & 0.80 & 0.84 \\
    \hline
  \end{tabular}
  \caption{Prior and posterior precision, recall and F-score of 3
    baselines (\leb and Empirical(-P)), 2 alien vote models (AV and
    IAV) and the IAV model in high-recall mode. The neural models
    outperform the empirical models and the alien vote models
    outperform the neural baseline.}
  \label{tab:f1}
\end{figure}

Table \ref{tab:f1} reveals a number of interesting model
characteristics.  First, all neural models achieve between 6\%-9\%
better posterior precision (Section \ref{subsec:datafscore}) than the
Empirical baseline. The IAV raw model is even able to achieve
comparable F-score to the empirical baseline even though it is
constrained to maintain 95\% precision.  When this constraint is
dropped (i.e., the IAV-HR model), the model dominates both the
Empirical and Empirical-P baselines by 12\% and 28\% F-score,
respectively. Note that the two baselines have no corresponding prior
performance because they rely entirely on the observed votes (i.e.,
they do not use side information).

The AV and IAV models outperform the neural baseline (\leb) in terms
of F-score. This supports our hypothesis regarding the advantages of
direct access to the alien votes during learning. Under the 5\% false
positive rate, the IAV model achieves the highest F-score of the
neural models. This might stem from its increased number of
parameters, but may also be the result of its information sharing
mechanism (Section \ref{subsec:iav}). We observe that the \leb
baseline has the highest precision, but generally exhibits
comparatively lower recall. We hypothesize that the \leb baseline is less
confident than the other models resulting in fewer predictions made by
$s(\cdot, \cdot)$. The \textsf{raw} and \textsf{prob} alien vote
representations produce slightly better models than the
\textsf{maj} representation, but no representation is dominant.

Finally, note that the precision of all neural models are close to
95\%, showing that the KB's precision can accurately controlled.

\subsection{Information Sharing and AVs}
\label{subsec:contrib-avs}
\begin{figure}[t!]
  \centering
  \begin{tabular}{ l c c c c c c }
    \hline
    & \multicolumn{3}{c}{Prior} & \multicolumn{3}{c}{Posterior}\\
    Model    & PRE  & REC  & F1   & PRE  & REC  & F1\\
    \hline
    AV raw & 0.96 & 0.49 & 0.65 & 0.96 & 0.59 & 0.73\\
    Zeroed & 0.97 & 0.37 & 0.53 & 0.97 & 0.52 & 0.68\\
    \hline
  \end{tabular}
  \caption{F-score of the AV raw and Zeroed models.  The Zeroed model
    has worse recall but high precision suggesting that
    alien votes help to increase model confidence.}
  \label{tab:zeroed}
\end{figure}

We are interested in understanding why the alien vote models perform
better than the \leb baseline. To do so, we conduct the following
experiment: first, we train the AV raw model; then, we create a
\emph{Zeroed} model from the trained AV raw model by artificially
converting all input alien vote vectors to zero vectors at test time.
Table \ref{tab:zeroed} compares the performance of the Zeroed and AV
raw models. While the two models have similar precision, the Zeroed
model has worse recall. This is because the Zeroed model is, overall,
less confident in its estimates (i.e., lower $\tau_{la}$). Lower
confidence leads to fewer predictions made by $s(\cdot, \cdot)$ (i.e.,
a higher number of ``No Prediction'' outputs), which in turn leads to
lower recall.  We conclude that the AV model estimates yes
rates effectively with side information alone and learns to use the
alien votes to boost confidence.

\begin{figure}[t!]
  \centering
  \centerline{\includegraphics[width=0.7\columnwidth]{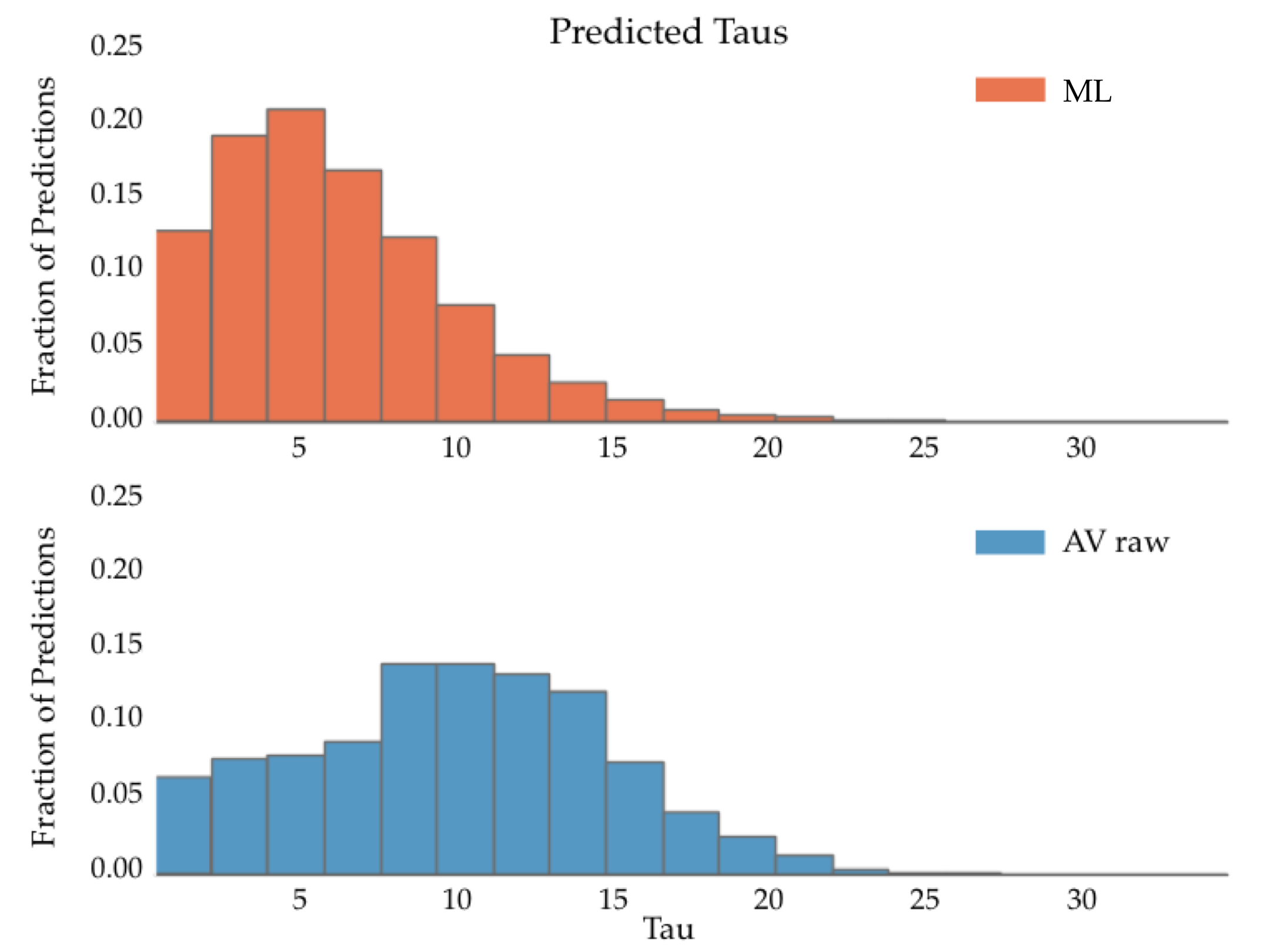}}
  \caption{Histograms of estimated $\tau_{la}$ (model confidence) for
    all location-attribute pairs in the test set.}
    \label{fig:taus}
\end{figure}

Figure \ref{fig:taus} shows histograms of $\tau_{la}$ estimated by the
\leb baseline and AV raw models. These histograms are evidence that
the \leb baseline tends to estimate lower values of $\tau_{la}$ than
the AV model. This supports our hypothesis that the \leb baseline's
relatively lower confidence is the cause of its lower F-score.

\subsection{Model Calibration}
\label{subsec:calibration}
For a more precise evaluation, we plot model calibration of the \leb
and AV raw models. To measure calibration, we first collect a test set
of all location-attribute pairs that received $n$ total votes. Next,
we select two integers, $Y$ and $N$--representing a number of yes and
no votes respectively--such that $Y + N = n$.  We use each model to
compute $Pr(Y | k, \mu_{la})$, i.e., the probability of observing $Y$
yes votes and $N$ no votes, for each location-attribute pair in the
test set. The locations are binned according to these probabilities.

\begin{figure*}[t!]
  \centering
  \begin{subfigure}[t]{0.49\textwidth}
    \centerline{\includegraphics[width=1.0\columnwidth]{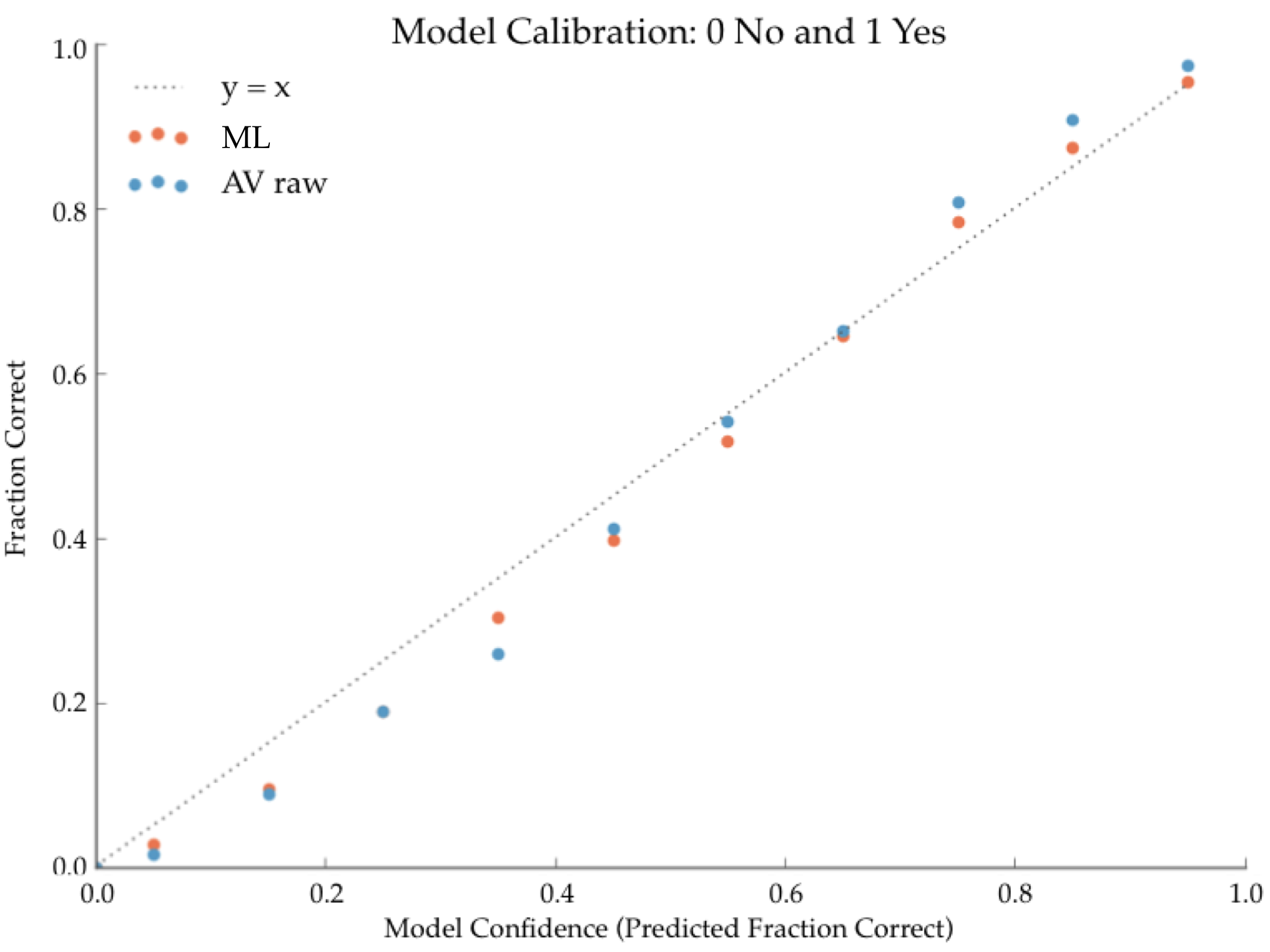}}
    \label{fig:cala}
  \end{subfigure}%
  \begin{subfigure}[t]{0.49\textwidth}
    \centerline{\includegraphics[width=1.0\columnwidth]{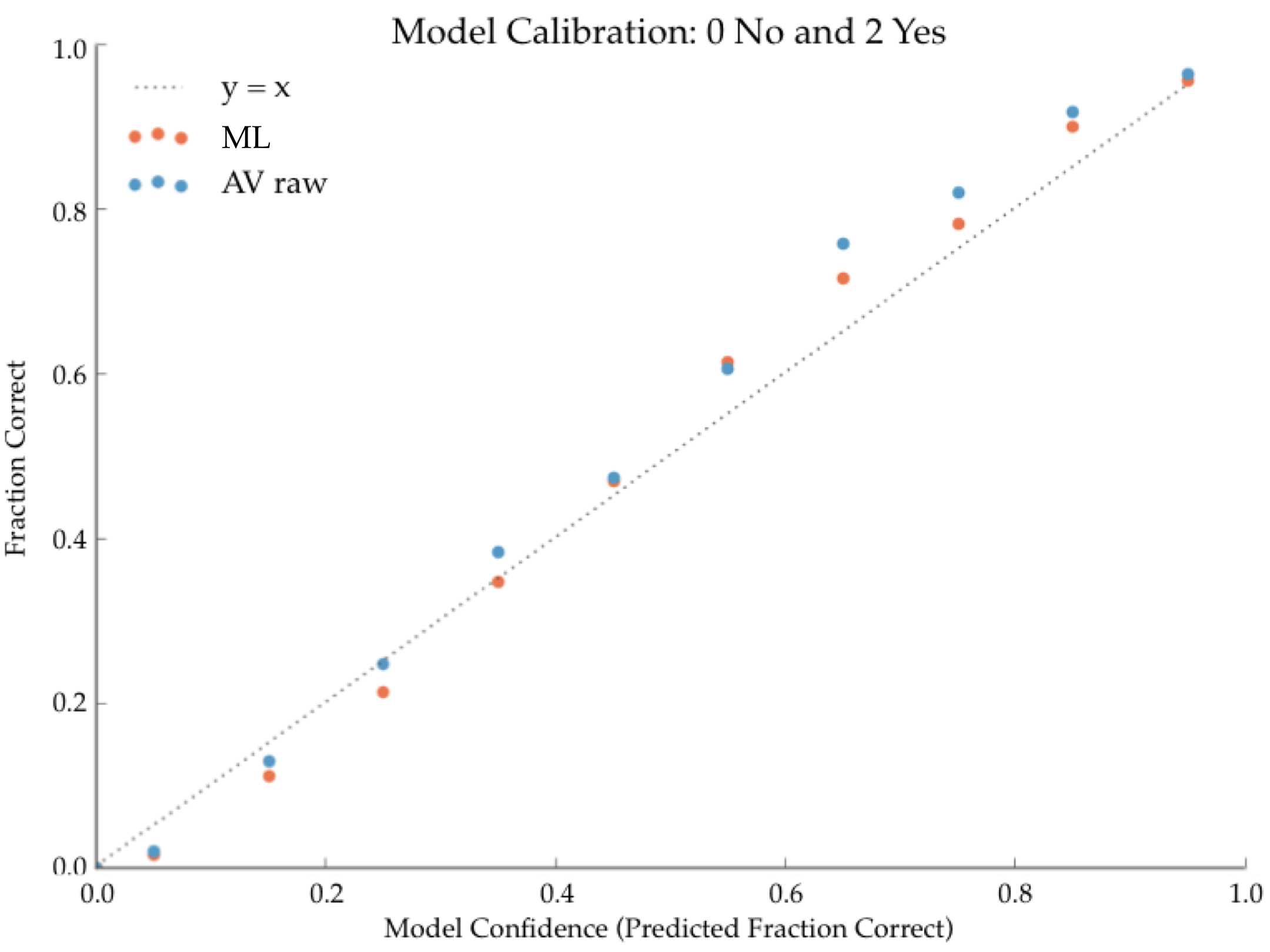}}
    \label{fig:av}
  \end{subfigure}%
  \caption{Calibration of the \leb and AV raw models for
    location-attribute pairs that received one and two total votes.}
  \label{fig:calibration}
\end{figure*}

We construct Q-Q plots showing the fraction of pairs in each bin for
which the observed votes were exactly $Y$ yes votes and $N$ no votes
(Figure \ref{fig:calibration}). For a well-calibrated model,
approximately $p$ (a fraction) of the items in the bin corresponding
to the probability $p$ should be classified correctly (i.e., have
exactly $Y$ yes and $N$ no votes). Since the predictions from both
models closely track the line $y=x$ (Figure \ref{fig:calibration}), we
conclude that our models are well-calibrated. We build Q-Q plots with
respect to 1 yes vote and 2 yes votes because these correspond to the
first and second moments of the prior yes rate distribution,
respectively. Q-Q plots corresponding to the other models reveal
similar trends with respect to calibration; they are elided for
brevity.

\section{Related Work}
\label{sec:related}
The literature on crowdsourcing for data collection and subsequent
model training is vast.  Most approaches collect multiple redundant
labelings for a set of tasks from a handful of crowd workers and then
infer the true task labels.  Even in cases where the tasks are
subjective, the true labels are considered to correspond to the
majority opinion \cite{meng2017subjective}.  Many of these methods
learn latent variable models of user expertise and task difficulty;
the learned models can be used for inferring the task labels
\cite{raykar2010learning, welinder2010multidimensional}. Some work
models both worker reputation and each item's label as a real-valued
random variable (in $[0,1]$) with a beta prior~\cite{de2015reliable}.
Like we do, other work develops beta-binomial models of the observed
labels \cite{carpenter2008multilevel}.  Unlike the prior art, we do
not explicitly model the crowd workers.  This is beneficial because it
does not require collecting a minimum number of labels per worker and
also protects worker anonymity. Whereas some previous work employs
expectation-maximization~\cite{whitehill2009whose}, variational
inference~\cite{liu2012variational}, Markov Chain Monte Carlo, or
variants of belief propagation \cite{karger2011iterative}, we estimate
parameters via back-propagation in neural networks. Some studies
develop intelligent routing of tasks to workers based on task
difficulty and user ability~\cite{karger2011iterative,
  ho2013adaptive}.  In our work, questions are routed to
geographically relevant users.

In this work, the model trained for  attribute prediction can be
characterized as a latent factors model. Previous work develops both
probabilistic and non-probabilistic factorization
models. Probabilistic matrix factorization techniques
\cite{salakhutdinov2007probabilistic} pose generative models of the
latent row and column representations and the observations. These
models can be extended to incorporate side information as in our set
up \cite{porteous2010bayesian, park2013hierarchical,
  liang2016factorization}. Much of the work on probabilistic matrix
factorization models assumes that the observations are corrupted with
Gaussian noise; this is not appropriate for our observed votes. More
similar to our work are studies of Poisson factorization, which naturally
models count data~\cite{gopalan2013scalable, gopalan2014content}.
While our work is not fully probabilistic, we do employ a generative
model of the observations.

There are a number of methods for learning parameters of factorization
models. For example, some methods leverage Bayesian Personalized
Ranking~\cite{rendle2009bpr, riedel2013relation} while others utilize
MCMC~\cite{liang2016factorization}.  Many modern methods, like ours,
perform learning via back-propagation in neural networks. These
approaches are trained end-to-end and incorporate side information
using various embedding techniques~\cite{wang2015collaborative,
  zheng2017joint, almahairi2015learning, elkahky2015multi}. Some work
demonstrates the benefits of multi-task learning for neural matrix
factorization~\cite{bansal2016ask}.

The incorporation of \emph{alien votes} (Section \ref{subsec:av}) is
similar to the \emph{CoFactor} model~\cite{liang2016factorization}.
The work on CoFactor optimizes an extension of Gaussian matrix
factorization and includes a column co-occurrence term. An analog to
this co-occurrence term in our framework is captured by attribute
relatedness (Section \ref{sec:empirical}) and made directly available
to our models via the alien votes. The AV architecture is loosely
inspired by wide and deep architectures~\cite{cheng2016wide}.
\section{Conclusion}
We study constructing a high precision KB of locations and their
subjective and factual attributes.  We probabilistically model the
latent yes rate of each location-attribute pair, rather than modeling
each pair as either True or False. Model confidence is explicitly
represented and used to control the KB's false positive rate. In
experiments, we demonstrate that our models: 1) are well-calibrated
and 2) that they outperform 1 neural and 2 empirical baselines.  The
experiments also reveal how different modes of information sharing
across attributes affect performance. While our experiments are
focused on the KB of locations and attributes that supports Google
Maps, our proposed framework is useful for constructing KBs with
tunable precision from unlabeled side information and noisy
categorical observations collected via crowdsourcing.

\newpage

\bibliographystyle{ACM-Reference-Format}
\bibliography{ms}

\end{document}